\documentclass[
hf,
]{ceurart}

\usepackage{listings}
\usepackage{amsmath, amssymb, amsfonts,enumitem}
\usepackage{algorithmic}
\usepackage{graphics, graphicx, subcaption, multirow, hhline}
\usepackage{xcolor}

\begin{document}

\copyrightyear{2022}
\copyrightclause{Copyright for this paper by its authors.
  Use permitted under Creative Commons License Attribution 4.0
  International (CC BY 4.0).}

\conference{In A. Martin, K. Hinkelmann, H.-G. Fill, A. Gerber, D. Lenat, R. Stolle, F. van Harmelen (Eds.), 
Proceedings of the AAAI 2022 Spring Symposium on Machine Learning and Knowledge Engineering for Hybrid Intelligence (AAAI-MAKE 2022), 
Stanford University, Palo Alto, California, USA, March 21–23, 2022.}

\title{Collusion Detection in Team-Based Multiplayer Games}


\author[1]{Laura Greige}[email=lgreige@bu.edu,]
\address[1]{Department of Computer Science, Boston University, Boston MA}

\author[2]{Fernando De Mesentier Silva}[email=fdemesentiersilva@ea.com,]
\address[2]{EA Digital Platform, Electronic Arts, Redwood City CA}

\author[3]{Meredith Trotter}[email=meredith.v.trotter@gsk.com,]
\address[3]{Artifical Intelligence \& Machine Learning, GlaxoSmithKline, San Francisco CA}

\author[4]{Chris Lawrence}[email=chlawrence@respawn.com,]
\address[4]{Respawn Entertainment, Los Angeles CA}

\author[1,5,6]{Peter Chin}[email=spchin@bu.edu,]
\address[5]{Center of Mathematical Sciences and Applications, Harvard University, Cambridge MA}
\address[6]{Center for Brains, Minds and Machines, MIT, Cambridge MA}

\author[2]{Dilip Varadarajan}[email=dvaradarajan@ea.com]

\begin{abstract}
In the context of competitive multiplayer games, collusion happens when two or more teams decide to collaborate towards a common goal, with the intention of gaining an unfair advantage from this cooperation. The task of identifying colluders from the player population is however infeasible to game designers due to the sheer size of the player population. In this paper, we propose a system that detects colluding behaviors in team-based multiplayer games and highlights the players that most likely exhibit colluding behaviors. The game designers then proceed to analyze a smaller subset of players and decide what action to take. For this reason, it is important and necessary to be extremely careful with false positives when automating the detection. The proposed method analyzes the players' social relationships paired with their in-game behavioral patterns and, using tools from graph theory, infers a feature set that allows us to detect and measure the degree of collusion exhibited by each pair of players from opposing teams. We then automate the detection using Isolation Forest, an unsupervised learning technique specialized in highlighting outliers, and show the performance and efficiency of our approach on two real datasets, each with over 170,000 unique players and over 100,000 different matches.
\end{abstract}


\maketitle

\section{Introduction}

Competition has always been a major element in games. It has become more relevant in recent years with the rise of esports and the games that are at its core. That becomes even more clear with the grown interest in battle royale games, where dozens of players take part in a match simultaneously, trying to be the last one standing. Some of the highest grossing games since 2017, when PlayerUnknown's Battlegrounds (PUBG) was first made available, are designed as battle royale. Games of such genre worth noting include Fortnite, Fall Guys: Ultimate Knockout, Apex Legends and Call of Duty: Warzone.

Along with competition, particularly on environments of large popularity, we usually find players who seek to cheat in order to get an unfair advantage. Cheating in video-games is accomplished in many forms, with the most common one being the use of software that provide a competitive edge. What is considered cheating in a specific game is defined in the user guidelines that players have to agree to before competing. Players are quick to express their discontent when they find themselves in the presence of unfair competition, and leaving cheaters unpunished negatively impacts the player base and quickly drives the majority of the community away from the game. In order to maintain a fair competitive environment, companies employ multiple different approaches to detect cheating. The most common approach includes allowing players to report players they suspect of cheating through an in-game interface. The game team then proceeds to review each player report and eventually takes action against confirmed cheaters if they deem it necessary, with the most common forms of punishments being short and long term banning of cheaters from competition.

When games are the target of common cheating methods, algorithms can be implemented to help identify them. Examples include the use of machine learning to identify what is commonly referred to as ``aimbot'', an external software or game hack that allows a user to automatically aim at adversaries with extreme efficiency. Algorithms are effective in identifying blunt, straight-forward cheating behaviors, but more specialized techniques need to be developed to deal with less transparent cheating approaches.

One more complex form of cheating is collusion. Collusion in multiplayer games can be described as players coordinating to purposely cooperate to gain an unfair advantage in a scenario in which they are posed as adversaries. There are different ways for a player to collude and one example includes communicating with an adversary to join forces and undermine other opposing players. This type of behavior involves analyzing multiple players, does not involve ``super-human'' actions, and therefore is harder to identify.

Since cheating is a strong offense in the context of a competitive game, and one that impacts other players negatively, it is natural that game designers would take strong action against the offenders. Strong actions in this scenario would involve banning a player from competing in the game, either temporarily or permanently. In many cases, this means blocking a user from accessing an account they have invested time and money in, either by purchasing the game or by making in-game transactions. For these reasons, it is necessary to be extremely careful with false positives and in guaranteeing that no user will be incorrectly banned from the game. Overall, failing to catch certain cheaters is still a more desirable outcome then banning a player that has been falsely labeled as a cheater.

The work we introduce in this paper is aimed at helping game designer with the task of identifying players that are colluding in their game. We present a novel approach to automate collusion detection in team-based multiplayer games and give proof-of-concept experimental results on two real-world datasets. Our proposed technique does not take any action on players. Instead, the purpose is to evaluate a player's gameplay history and behavioral pattern, and with our prior knowledge and assumption on colluding behaviors, we assign an anomaly score to each pair of opponents indicating the degree of collusion exhibited by the pair's respective teams. The game designers can then anlayze more thoroughly the group of players our technique flagged as potential colluders, and decide whether to take action on them or not. This would be a great improvement in their current workflow, where they either need to manually look through every player, which is unfeasible with the active players numbers in the millions, or only look at players that are brought to their attention by the community, which would allow many players to continue to cheat and ruin the experience of others, which can lead to higher churn rate.

\section{Related Work}

Collusion detection has been an exciting area of research and has been greatly analyzed in different fields including card games \cite{bib:automating-seq-games, bib:colluders-poker}, online reputation systems and auctions \cite{bib:online-rep-systems, bib:government-procurement-auctions}, and multiplayer single-winner games \cite{bib:eliciting-collusion-features, bib:towards-accurate-collusion-detection, bib:impact-of-collusion, bib:detecting-colluding-subset}. However, most approaches to detecting collusion use supervised learning with the assumption that large datasets containing both confirmed colluding and normal behaviors are available. In practice, such training sets are not always available, particularly in online video games. Additionally, these models are based on a colluder's past actions and behavior causing potential new forms of collusion to go undetected.

In our work, we make no assumption on the players' behaviors and turn to unsupervised learning algorithms specialized in separating colluding behaviors from the rest. Our paper also extends the research on collusion detection in online video games to team-based multiplayer games, in order to prevent players from collaborating with other teams of players during a match. To the best of our knowledge, this is the first work that focuses on detecting collusion in a team-based multiplayer game. The rest of the paper is structured as follows. In the next section, we define the problem of collusion in games and introduce the definition and notations necessary to understand the paper. Section \ref{sec:method} introduces the game environment and the dataset used in our experiments. This section also describes the general characterization of the feature set selected differentiating colluding teams from the rest and describes our approach for detecting colluders. Section \ref{sec:results} discusses our general findings and shows the efficiency of our approach on real datasets, followed by concluding remarks in Section \ref{sec:conclusion}.

\section{Method}
\label{sec:method}

There is a characteristic of the problem at hand that guides our technique. There is no current way to detect colluders, other than having game designers manually check an individual's gameplay data. Only a handful of colluders have been identified so far (less than 100), meaning there is no or not enough labeled data to be used and by consequence, we cannot rely on supervised learning.

In the next sections, we define cross-team collusion in games based on an individual's social relationships and gameplay data. The game platform allows users to befriend one another, allowing them to stay up to date with their friends' progression, share their own progression and chat about gameplay strategies. Therefore, social relationships are defined by an individual’s external connections on the game platform and by its gameplay history, particularly, its in-game teammate and opponent relationships. Using a player’s external and in-game social relationships paired with its in-game behavioral patterns, we select a feature set that allows us to differentiate colluding teams from the rest. Hence, the problem comes down to detecting anomalous observations in our datasets. An anomaly is defined as any data point that differs from the norm, in our case, any team behavior and interaction that differ from the rest. Colluding has a negative impact for other, non-colluding, users in online video games and draws bad reputation to video game companies through social media posts and campaigns. The goal of this paper is to automate the detection of colluding teams in team-based multiplayer online game. With large amounts of data at hand, data mining and AI techniques can help game designers tackle the problem of collusion detection where manual analysis and detection may be impossible. We propose a system that detects and measures the degree of collusion exhibited by all pairs of opponents, based on their behavioral patterns as well as the game designers' knowledge regarding colluding behaviors in team-based multiplayer games. Note that our system does not take any action on the detected outliers. It only identifies a set of suspected colluding teams that will still require further human investigation, giving game designers some prior assumption on who may be colluding.

In the remainder of the paper, we consider a set of $n$ players and $m$ teams, such that $m < n$. We use real-valued vectors to describe individual player and team features such that one match opposes 20 teams with at least 2 players each, and is described by the combination of feature sets of all teams involved in that match.

\subsection{Game Environment}

Consider a game where teams of two players or more take part in a match simultaneously and compete in an open environment that is known by all players. The goal of the game is to be the last team standing. Teams are eliminated as the game progresses, either by other teams, or by standing outside the safe area. The safe area is the space in the environment where normal conditions apply, standing anywhere outside of it pressures players, as they are eliminated for standing out of the safe area for too long. As the match progresses, the safe area is purposely reduced over time, to force players into a more crowded space, leading to climatic moments at the end of the game, when the safe area is small and teams are forced into conflict. The game begins with teams starting at different locations. Each team chooses the location they will start in, but their starting position is not known by the other teams. Gameplay revolves around the teams exploring the environment. As they go through the map, they are constantly keeping watch for other players, as learning the position of other teams is crucial for having an advantage over them. Another important element are items that are scattered through the map, collecting them gives the teams an edge, making it easier to eliminate other players. It is important to note that opponents have no way of verbally communicating through the game. Each team is assigned a final rank placement at the end of a match. When all players in a team are eliminated, their placement is equal to the number of teams remaining in the game plus one, with any number of players in that team yet to be eliminated. In other words, if there are $n$ teams in a match, then the first team to be eliminated is ranked $n$-th, the second team to be eliminated is ranked $(n-1)$-th and so on. Match events logs are captured during gameplay and stored at the end for processing. Events provide information regarding the game state, such as players locations and actions taken throughout the match, as well as information regarding the teams (e.g. teammates, team rank placement, etc).

\subsection{Data Collection}

Our analysis is based on data from team-based tournament modes of an online team-based multiplayer game, including cases of colluders confirmed by the game designers. We use data from 113,060 unique matches and 173,603 unique players that have played over 3 matches over the span of 3 days. We strengthen our findings by using data from 44,097 unique matches and 128,342 unique players that have played over 3 matches over the span of 2 days, at a different time range, that includes a handful of confirmed cases of colluders. Player data is collected in accordance with applicable privacy policies and collected based on players’ privacy settings and preferences. In what follows, when we refer to the game or the environment, we imply that all previously mentioned conditions apply. 

\subsection{Data \& Knowledge}

One of the most important factors in hybrid AI systems is knowledge. Knowledge-based AI approaches can be characterized by observations and relationships inferred by human experts and their past experiences. Machine learning approaches, on the other hand, implicitly derive these relationships from the data at hand. The concept of hybrid AI systems aims at using the complementary strengths of human intelligence and expertise, combined with data-driven learning, to collectively achieve better results and improve the efficiency of machine learning processes.

We construct individual players' feature sets from their in-game attributes, as described in the game logs. We focus on attributes such as match experience (e.g. number of matches a player has participated in, final team rank placement) and play style (e.g. player's starting position in each match played). The choice of features used in our technique comes from a combination of the game designers' expert knowledge on player behaviors and what differs colluding players from non-colluding ones, but also exploration and experimentation with the feature set. Initial investigations into colluding behaviors were initiated by instances of user complaints and social media posts. As an example, video evidences showed a set of players from opposing teams deliberately choosing not to fight each other despite their close proximity, teaming up against other teams to ensure first and second place victories in ranked matches. Hence, when constructing the team feature set, we look at pairwise attributes such as the number of matches both teams participated in, their landing proximity to one another and their final team rank difference at the end of each match played. Using individual player and team features, we selected the following feature set to infer information differentiating colluding teams from the rest : landing proximity, final team rank placement, acquaintance, number of matches and number of consecutive matches.

\subsection{Feature Selection}

\begin{table}[h]
\def\arraystretch{1.1}
\centering
\resizebox{\columnwidth}{!}{%
\begin{tabular}{|l||c|c||c|c|}\hhline{~----}
\multicolumn{1}{c|}{} &  \multicolumn{2}{c|}{Dataset 1} & \multicolumn{2}{c|}{Dataset 2}\\\hhline{~----}
\multicolumn{1}{c|}{} & {\textbf{Teammates}} & \multicolumn{1}{c|}{\textbf{Opponents}} & {\textbf{Teammates}} & \textbf{Opponents}\\\hhline{-====}
Number of Pairs & 171,794 & 347,890 & 95,893 & 65,744\\\hhline{-----}
Acquaintances & \multicolumn{2}{c||}{469} & \multicolumn{2}{c|}{143}\\\hhline{=====}
Average \# of matches played & 14.4 & 4.6 & 9.2 & 4.5\\\hhline{-----}
Maximum \# of matches played & 217 & 35 & 95 & 38\\\hhline{=====}
Average distance & 1,886.35 & 31,433.17 & 2,115.45 & 31,694.38\\\hhline{-----}
Average rank difference & N/A & 5.19 & N/A & 5.36\\\hhline{=====}
Appeared in 3+ consecutive matches & 44,229 & 22,472 & 23,056 & 4,196\\\hhline{-----}
\end{tabular}}
\caption{Gameplay statistics for Dataset 1 and Dataset 2. Numbers are coherent over both datasets, in particular the average distances between teammate and opponent landing positions and the average rank differences. Dataset 1 was queried over the span of 3 days whereas Dataset 2 was queried over the span of 2 days, which explains the lower number of opponents that have appeared in the same set of 3 or more matches.}
\label{tab:stats-full}
\end{table}

Initial statistics are reported in table [\ref{tab:stats-full}] for both datasets. As a reminder, both datasets include data from pairs of players that have played over 3 games in the span of 3 days for dataset 1 and 2 days for dataset 2. The probability that two teammates appear in the same set of matches is naturally higher than that of two opponents, which explains the significantly higher number of pairs of teammates in both datasets. Moreover, it is highly unlikely to find a pair of players that not only has played over 3 games as teammates but also appeared in at least 3 or more matches as opponents, more so over such a short period of time. However, we find 469 pairs in Dataset 1 and 143 in Dataset 2 such pairs and include these values in the table under acquaintances. As stated in the previous section, we are interested in players' proximity to one another, hence we look at the average distance between each pair of players' starting positions. Values are coherent over both datasets where the average distances between all pairs of teammates are 1,883 for Dataset 1 and 2,125 game units for Dataset 2, and the average distances between all pairs of opponents are 31,928 for Dataset 1 and 31,112 game units for Dataset 2. Similarly, the average rank difference between two pairs of opponents is also coherent over both datasets with a 5.27 average for Dataset 1 and a 5.28 average for Dataset 2.

\subsubsection{Proximity}

\begin{figure}[h]
\centering
\rotatebox[origin=c]{90}{Dataset 2\hspace{65pt}Dataset 1}\hfill
\begin{subfigure}[t]{0.31\textwidth}
    \includegraphics[width=\textwidth]
    {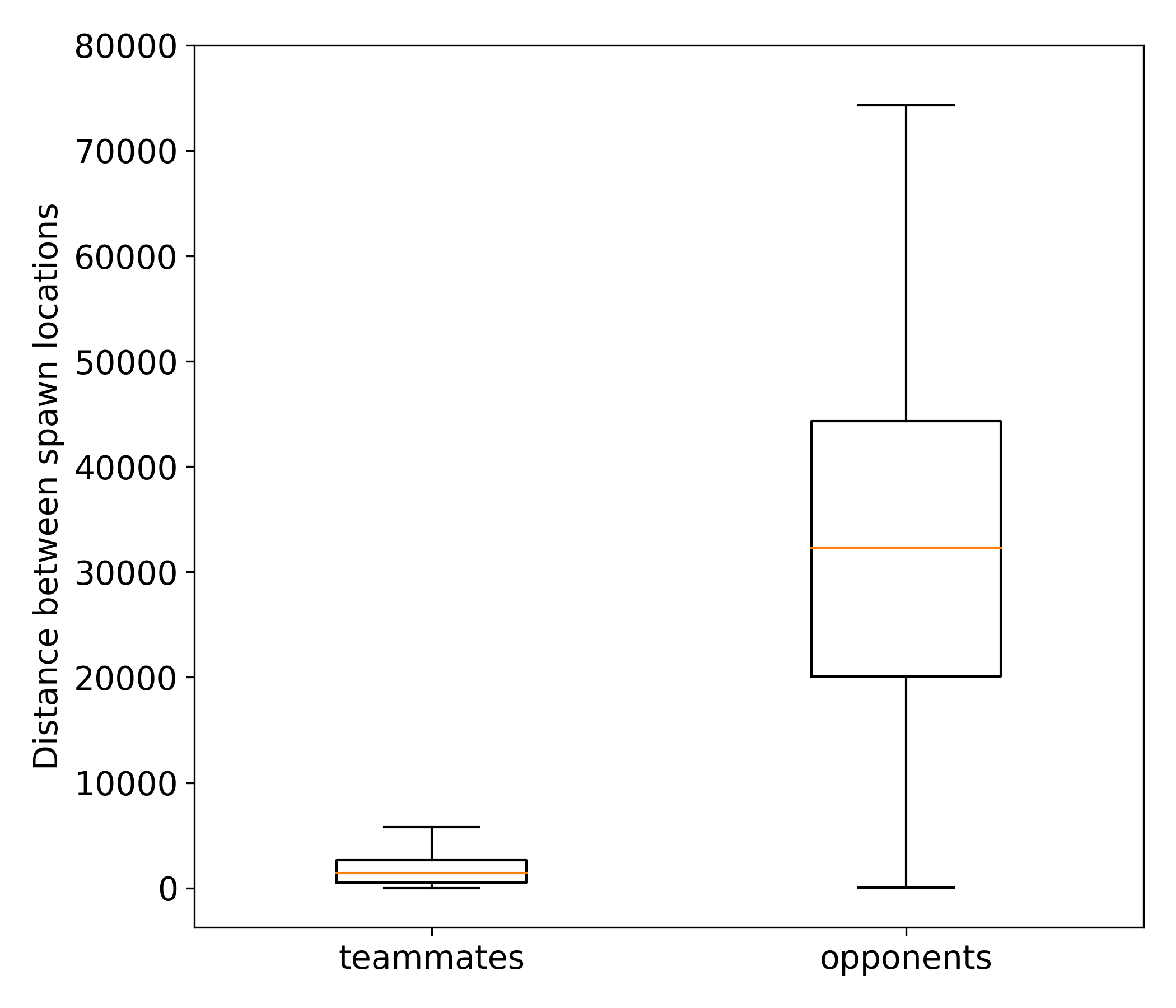}
    \includegraphics[width=\textwidth]
    {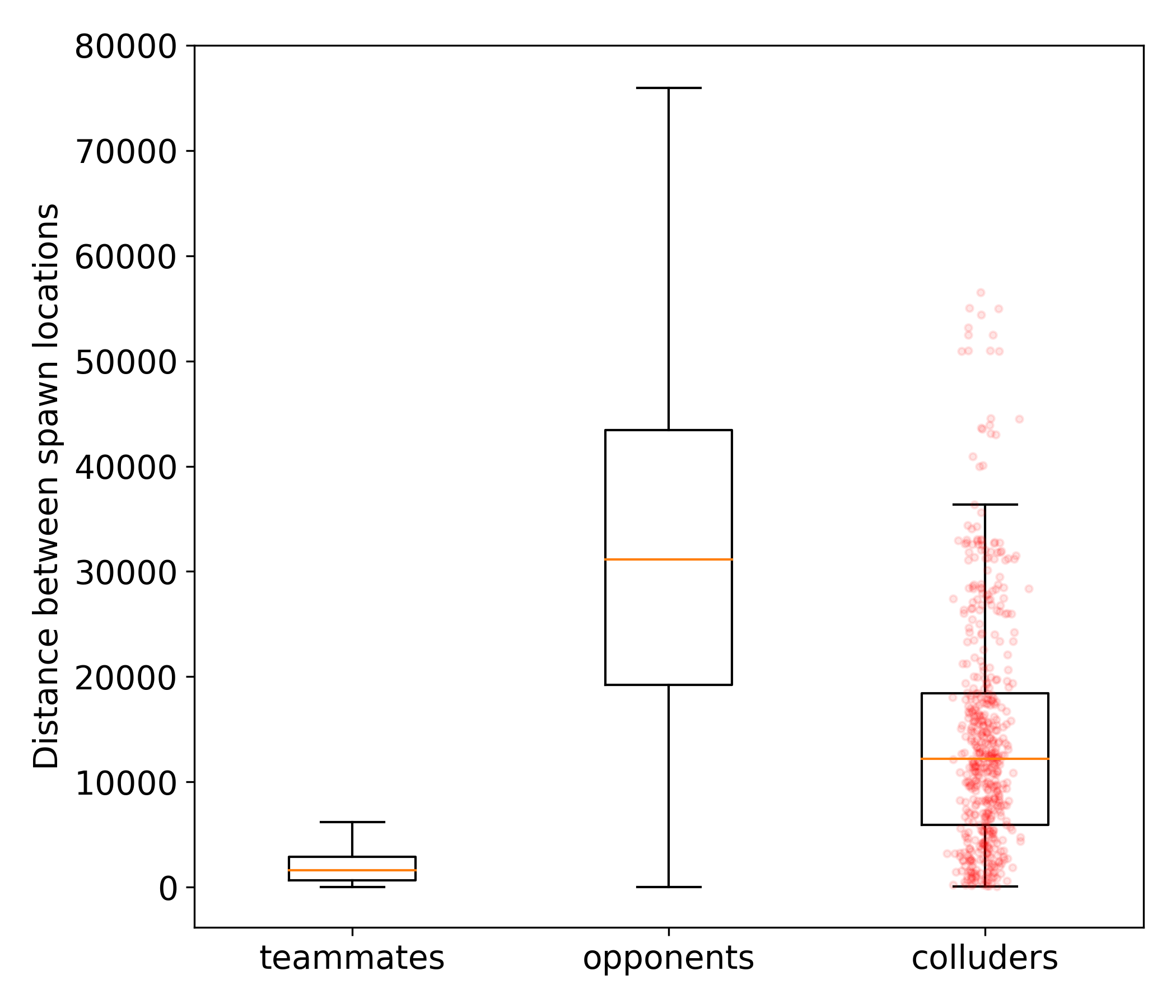}
    \caption{Distance distribution over all matches played for each pair of teammates and opponents. For Dataset 2, we add the distance distribution for pairs of confirmed colluders in red.}
    \label{fig:proximity-distribution}
\end{subfigure}\hfill
\begin{subfigure}[t]{0.31\textwidth}
    \includegraphics[width=\textwidth]  
    {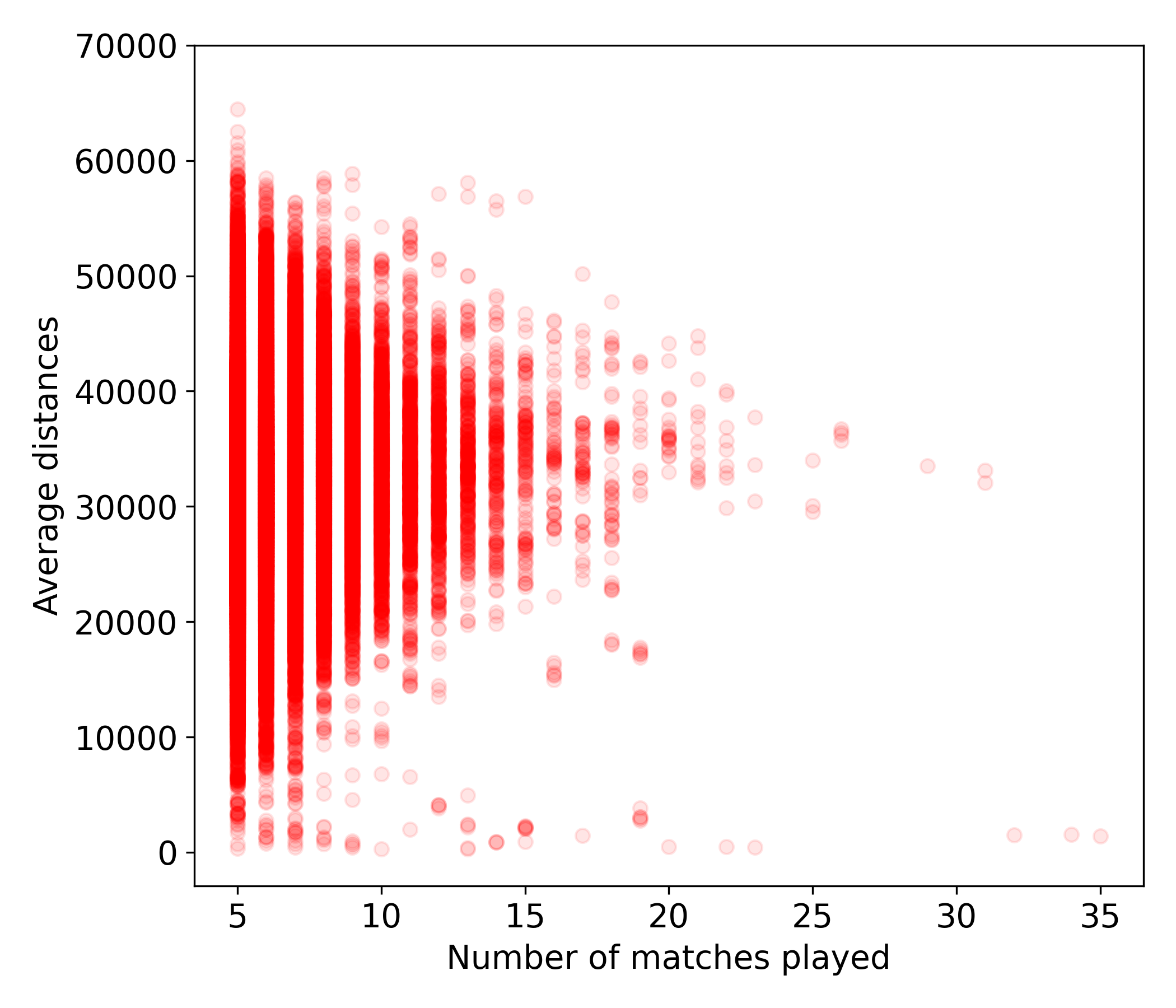}
    \includegraphics[width=\textwidth]
    {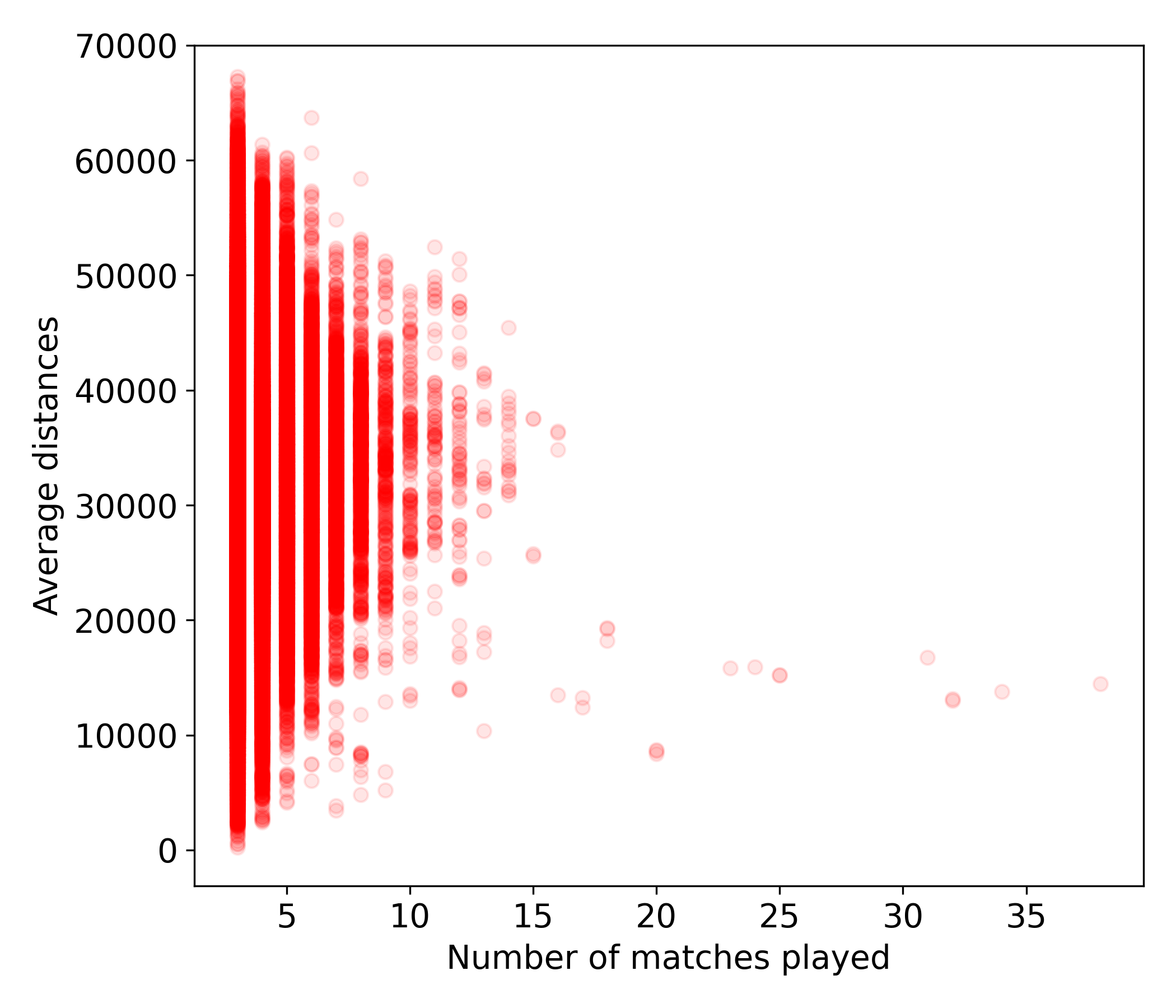}
    \caption{Average distance between each pair of opponents from Dataset 1 and 2, averaged over the set of matches both opponents participated in.}
    \label{fig:proximity}
\end{subfigure}\hfill
\begin{subfigure}[t]{0.31\textwidth}
    \includegraphics[width=\textwidth]  
    {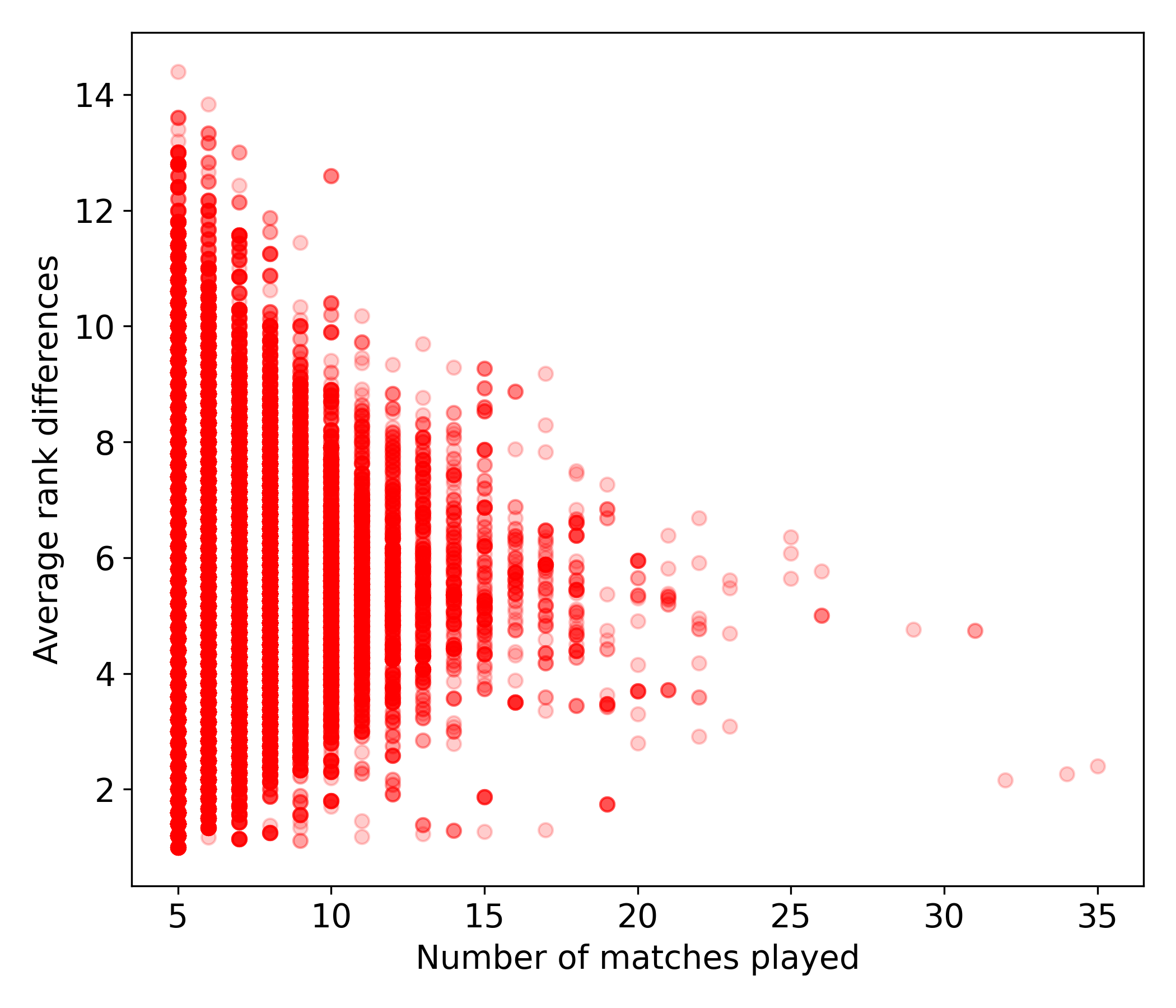}
    \includegraphics[width=\textwidth]
    {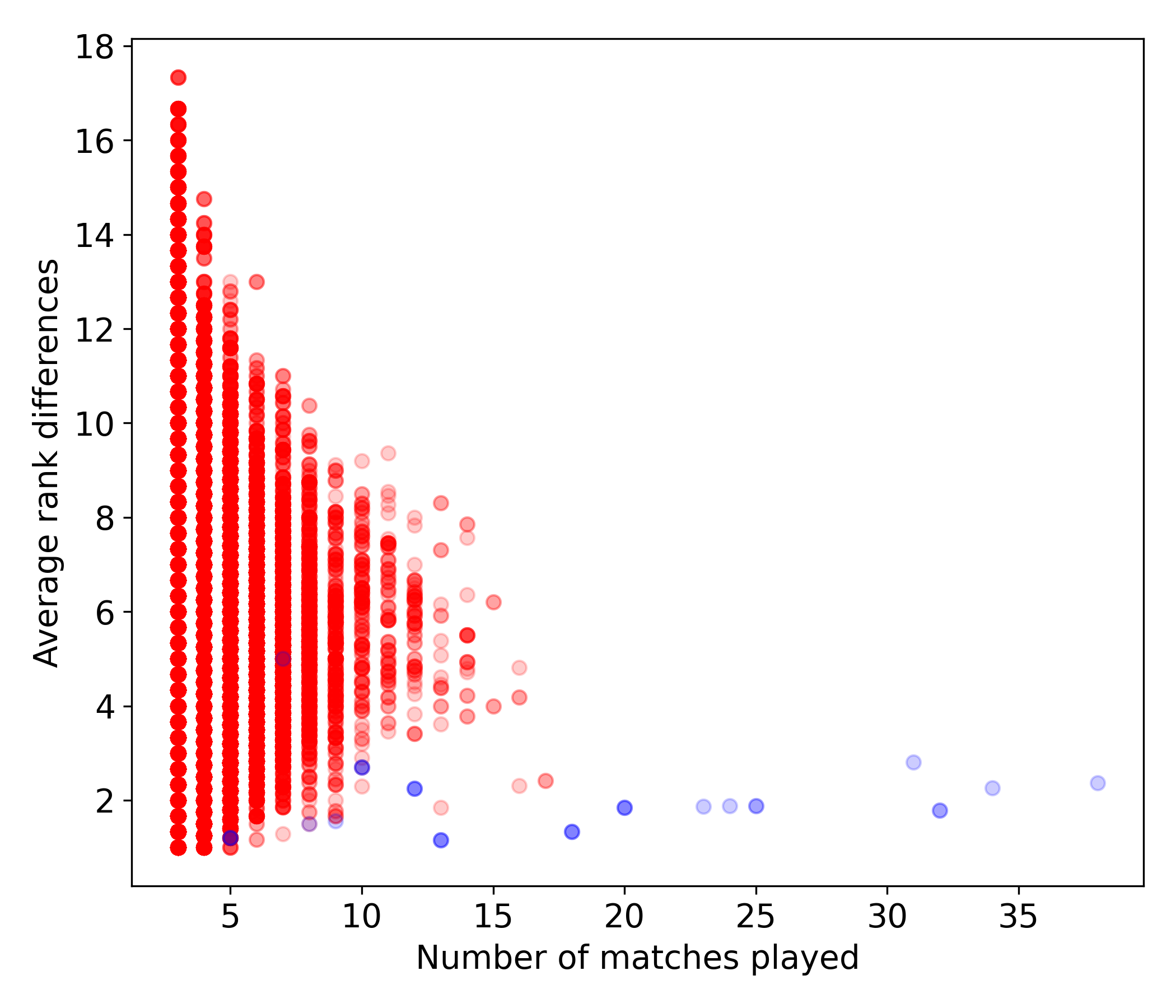}
    \caption{Average rank difference for each pair of opponents from Dataset 1 and 2. For Dataset 2, we highlight the pairs of confirmed colluders in blue.}
    \label{fig:rank-placement}
\end{subfigure}
\caption{Teammate and opponent distance distribution, average distances and average rank difference between opponents plotted in Figures \ref{fig:proximity-distribution}, \ref{fig:proximity} and \ref{fig:rank-placement} respectively. The top figures represent players from Dataset 1, while the bottom ones represent players from Dataset 2. Each dot in Figures \ref{fig:proximity} and \ref{fig:rank-placement} represents a pair of opponents.}
\end{figure}

For each pair of teammates and opponents, we calculate the average distance between players' initial positions in the game environment over all matches both players participated in. We plot the distribution of the average distances for all pairs of teammates and for all pairs of opponents from Dataset 1 and 2 in Figure \ref{fig:proximity-distribution}. We observe that the majority of teammates were in close proximity at the start of each match, with some exceptions nearing an average distance of 10,000 game units. In contrast, the average distance between two opponents varies greatly, ranging from near 0 to over 70,000 game units. In Figure \ref{fig:proximity}, we plot the difference between each pair of opponents' starting positions averaged over all matches both players participated in. With over 10 match appearances, certain pairs of opponents were as close as teammates would be.

Colluding teams have an incentive to remain close to one another, which begins by pre-planning map location starting points. Two opponents being in close proximity at the start of a game does not necessarily indicate collusion, but the together with the other features, especially when analysing a pair's proximity averaged over multiple matches, is an important indicator.

\subsubsection{Rank Placement}

There are a number of reasons why users would collude in games but ultimately, they collude to increase their chance of winning matches and become top-tier players, joining the highest ranks. These players have an incentive to collaborate with other top-tier players in order to aid each other in keeping their respective status and tier ranks. For this reason, we believe that colluding teams are more likely to be closely ranked at the end of each match. For each pair of opponents, we look at their team rank placement difference over all matches both players participated in as opponents. In Figure \ref{fig:rank-placement}, we plot these values for each pair of opponents. We also highlight the confirmed cases of colluders from Dataset 2 in blue.

Suppose two random teams finish one rank apart in a single match that involved 20 different teams. Then the probability of this event occurring is equal to $(2 \cdot 19 \cdot 18!)/20! = 0.1$. Therefore, the probability that two teams finish one rank apart $k$ times in $n$ matches would be equal to the following:
\begin{align}
{n \choose k} (0.1)^k(0.9)^{(n-k)}
\end{align}

\noindent Now suppose two random teams finish one rank apart in a single match but also rank in the top 10 spots out of 20 spots. The probability of this event occurring is equal to $(2 \cdot 9 \cdot 18!)/20! = 0.047$. Therefore, the probability that the same two teams finish in the top 10 spots and finishing one rank apart in $n$ different matches is equal to the following:

\begin{align}
{n \choose k} (0.047)^k(0.953)^{(n-k)}
\end{align}

To give a better idea, suppose both teams finished in the top 10 spots and one rank apart in 3 matches out of 5. Then the probability of this event occurring is equal to 0.00094. Despite the very low probability of this event occurring, we find 6,771 pairs of opponents in Dataset 1 and 26,370 pairs of opponents in Dataset 2 ranked in the top 10 out of 20 teams, with an average rank difference less than or equal to 2 out of over 110,000 and 290,000 unique pairs of opponents respectively. Note that these values represent pairs of opponents on different teams and not individual teams. 

\subsubsection{Acquaintance}

To coordinate a collusion in a game, we believe that  players involved are acquaintances. For this reason, we look at a player's social relationships. Social relationships are defined by the connection between two individuals on the game platform which allows users to befriend one another, and by their gameplay history, particularly, their in-game teammate and opponent relationships.

We find 469 pairs of players that have played both as teammates and as opponents in at least 3 games in Dataset 1 and 143 pairs in Dataset 2. There is no sure way to face a predetermined opponent in the game and players are matched at random with players of similar skill levels. Therefore, it is generally highly unlikely for two players to end up as opponents in multiple matches unless some collusion such as communicating outside the game to coordinate playing at the same time and on the same game server, occurred prior to the matches, increasing their chance of being assigned to the same set of matches.

\subsubsection{Number of Matches and Consecutive Ones}

Finally, we look at the number of matches and consecutive matches each pair of opponents has participated in. We assume that colluding teams are more likely to have participated in multiple matches with the same teammates and/or opponents. The higher the number of matches two players appeared in, the more likely it is that these players agreed on playing the same set of matches. We are also more interested in players that constantly collude, ruining the user experience for other players. We are not so much concerned about short-term and temporary colluders, as they only temporarily disrupt the experience for the rest of the players. Therefore, in what follows, we only consider pairs of players that have participated in at least 5 matches together for Dataset 1 and in at least 3 matches for Dataset 2. Finally, we also look at the number of consecutive matches two opposing teams participated in as a higher number of consecutive matches gives more insight on whether appearing in the same set of matches was premeditated.

Each individual feature is naturally not enough to determine whether two or more teams have been colluding and therefore, it is important to consider a combination of the different features to better differentiate colluding behaviors from the norm.

In this section, we study the relationship between teammates and opponents using social network analysis and explore the efficiency of our model on real datasets while showing its performance on confirmed cases of collusion.

\section{Experimental Results}
\label{sec:results}

\subsection{Social Network Analysis}

So far, we have analyzed the behavior of pairs of teammates and pairs of opponents with a focus on their gameplay history. We also aim to understand the connections between players from different teams in different matches. Social network analysis (SNA) is one tool that could potentially enhance our understanding of the relations that develop between different teams throughout the matches played, and that could indicate collusive behaviors. SNA uses concepts from graph theory to calculate metrics representing the nodes, the connections or the network itself such as the distance to other individuals in the network or the number of interactions with other individuals. These metrics are important in quantifying interactions between players or teams of players as well as identifying potential clusters in the network. Moreover, network analysis provides a visual model for analyzing and evaluating teammate and opponent relationships.

\begin{figure}[h]
\begin{center}
\subcaptionbox{Dataset 1\label{sub:sn-1}}
{\includegraphics[width=.49\textwidth]{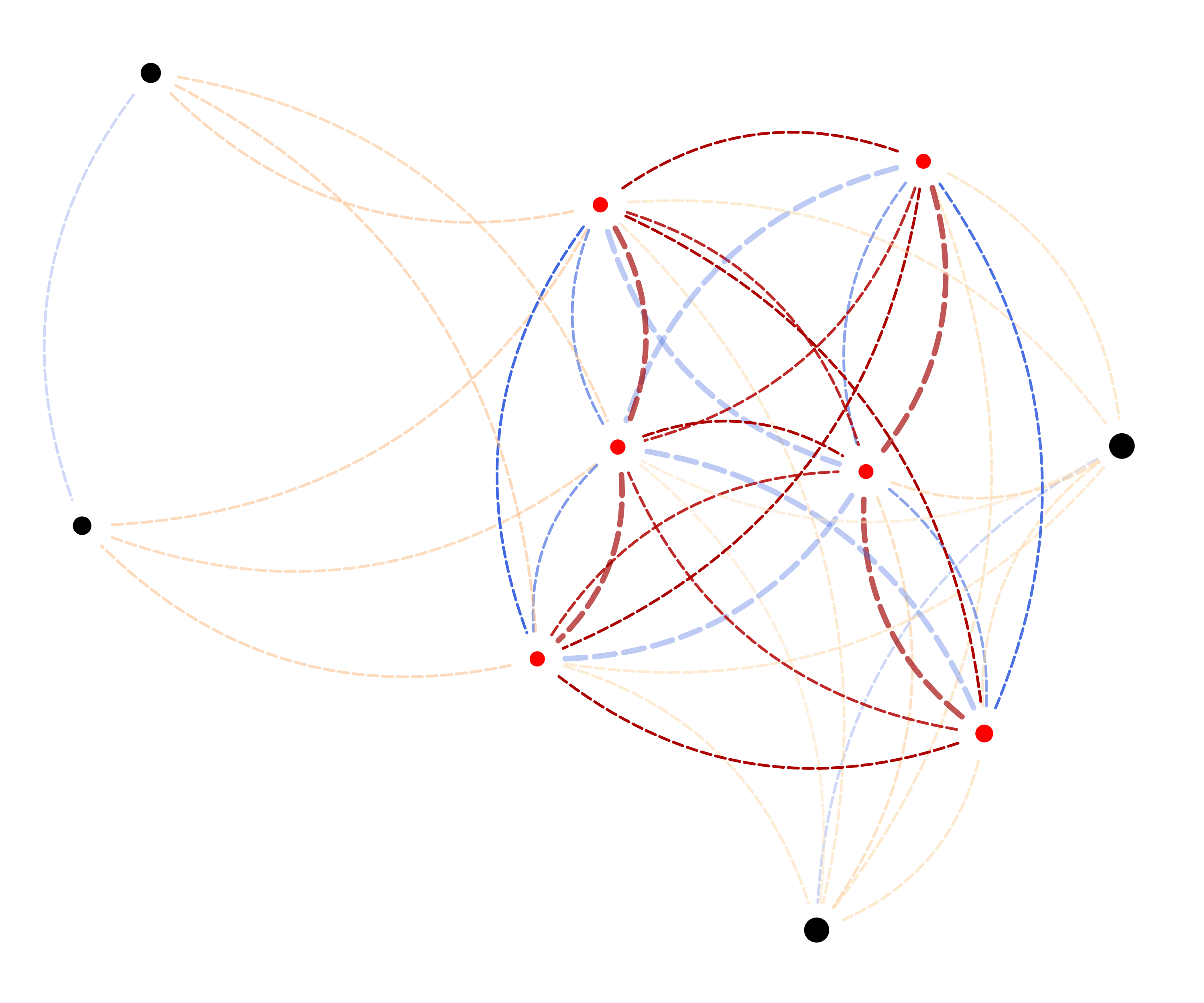}}
\subcaptionbox{Dataset 2\label{sub:sn-2}}
{\includegraphics[width=.49\textwidth]{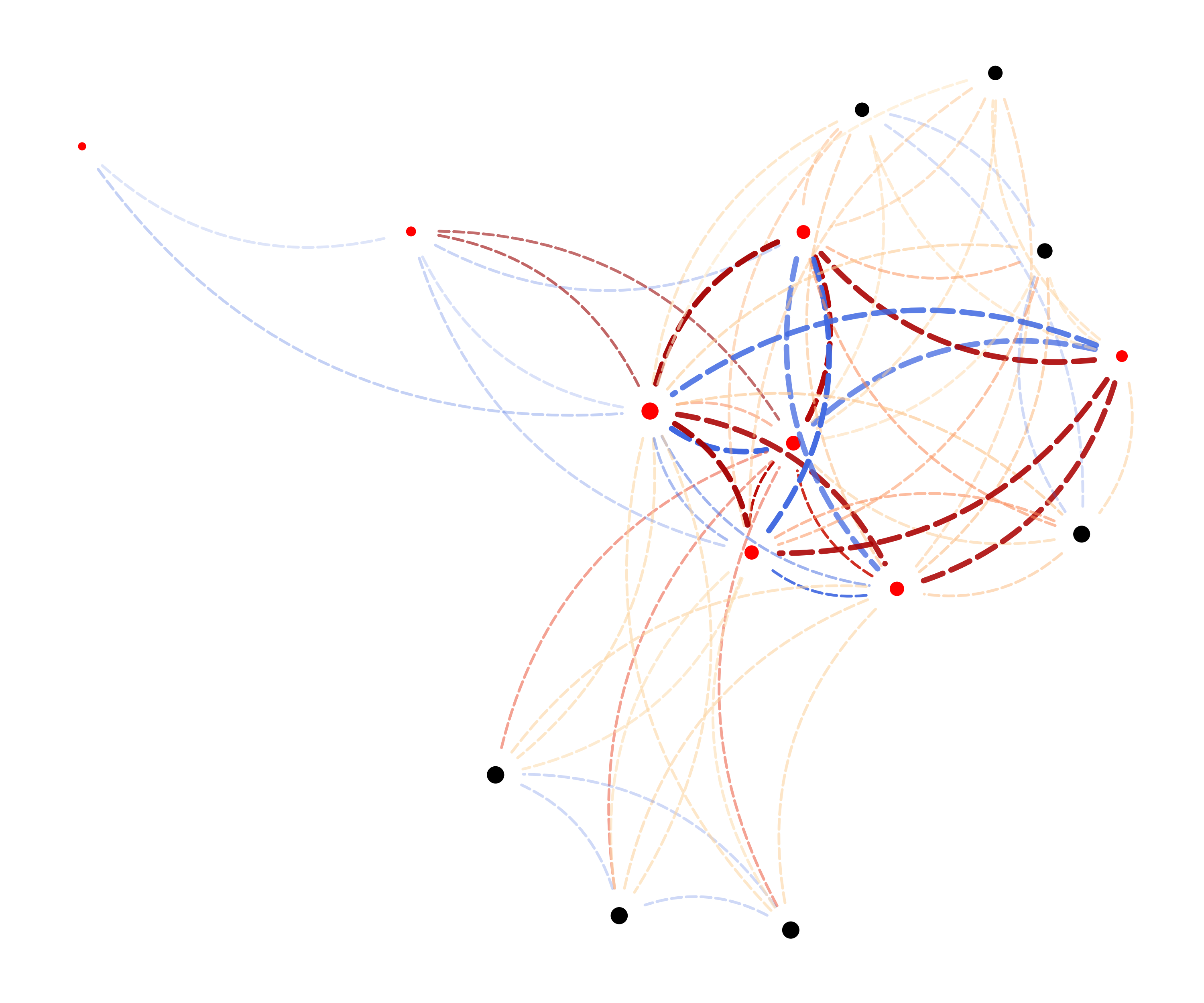}}
\caption{Social network illustrations of a set of suspected colluders from Dataset 1 in Figure \ref{sub:sn-1} and a set of confirmed colluders from Dataset 2 in Figure \ref{sub:sn-1} represented by red nodes. Players from each set are tightly connected whether by close final team rank placement (red colored edges), or by appearing in a larger number of matches (darker opacity), or both, as opposed to the rest of the pairs of opponents in each social network. The biggest difference between both sets is the fact that the confirmed colluders from Datset 2 appeared in more matches (occasionally consecutive matches) as teammates (blue edges), as opposed to the suspected colluders from Dataset 1.}
\label{fig:social-network}
\end{center}
\end{figure}

For clarity purposes, we only plot edges between players that have appeared in over 3 matches together. Each node in the social network represents an individual player and the larger the node size, the more matches that user has played. Teammate relationships are represented by blue edges while opponent relationships are represented by edges of colors ranging from yellow to maroon such that the darker the color of the edge is, the closer the two players are in terms of rank placements averaged over all matches appeared in together. The darker the opacity of an edge is, the higher the number of games the two users have appeared in. Similarly, the thicker the edge is, the higher the number of consecutive games the two users have appeared in. A number of clusters get formed and two examples are shown in Figure \ref{fig:social-network}. Figure \ref{sub:sn-1} illustrates an example of a set of suspicious players as defined in the previous section, while Figure \ref{sub:sn-2} illustrates an example of a set of confirmed colluders, all represented by red nodes. We notice that these players are strongly connected in the social networks and each pair of players has either played as teammates, as opponents or as both, suggesting pre-established acquaintance. As stated earlier, the thicker and the darker the edge, the more significant the interaction was between both players. We obtain a number of different clusters, not shown in this paper, that exhibit similar behavior.

\subsection{Automating Detection}

Isolation Forest \cite{bib:isolation-forests} is an unsupervised learning algorithm used to detect outliers and to identify anomalies instead of normal observations, that is to identify rare events that deviate from the norm. Isolation Forests (IF) are tree-based algorithms built around the theory of decision trees and random forests. IF separates the data into two parts by randomly selecting a feature from the given feature set and then randomly selecting a split value between the minimum and maximum values of the selected feature. The random and recursive partition of data is represented as a tree such that the end of the tree is reached once each data point is isolated. Outliers need fewer random partitions to be isolated from the rest of the dataset, therefore the outliers will be the data points which have a shorter average path length from the root node on the isolation tree. In what follows, we choose 100 estimators with 1000 samples where each base estimator is trained on the features mentioned in the previous section. We do not have an accurate estimation of the number of actual colluders present in our dataset, but, for comparison, game designers have a rough estimate that, in  the game, there are 500 unique colluders every month with roughly 2000 user reports of others being suspicious of collusion behavior and dataset 1 covers a span of 3 days, while dataset 2 covers a span of 2 days. Given a dataset of players and matches, combined with individual player and team features, each pair of opponents is assigned an anomaly a score indicating the degree of collusion exhibited by the pair's respective teams.

\begin{figure}[h]
\centering
\begin{subfigure}[t]{0.49\textwidth}
\includegraphics[width=\textwidth]{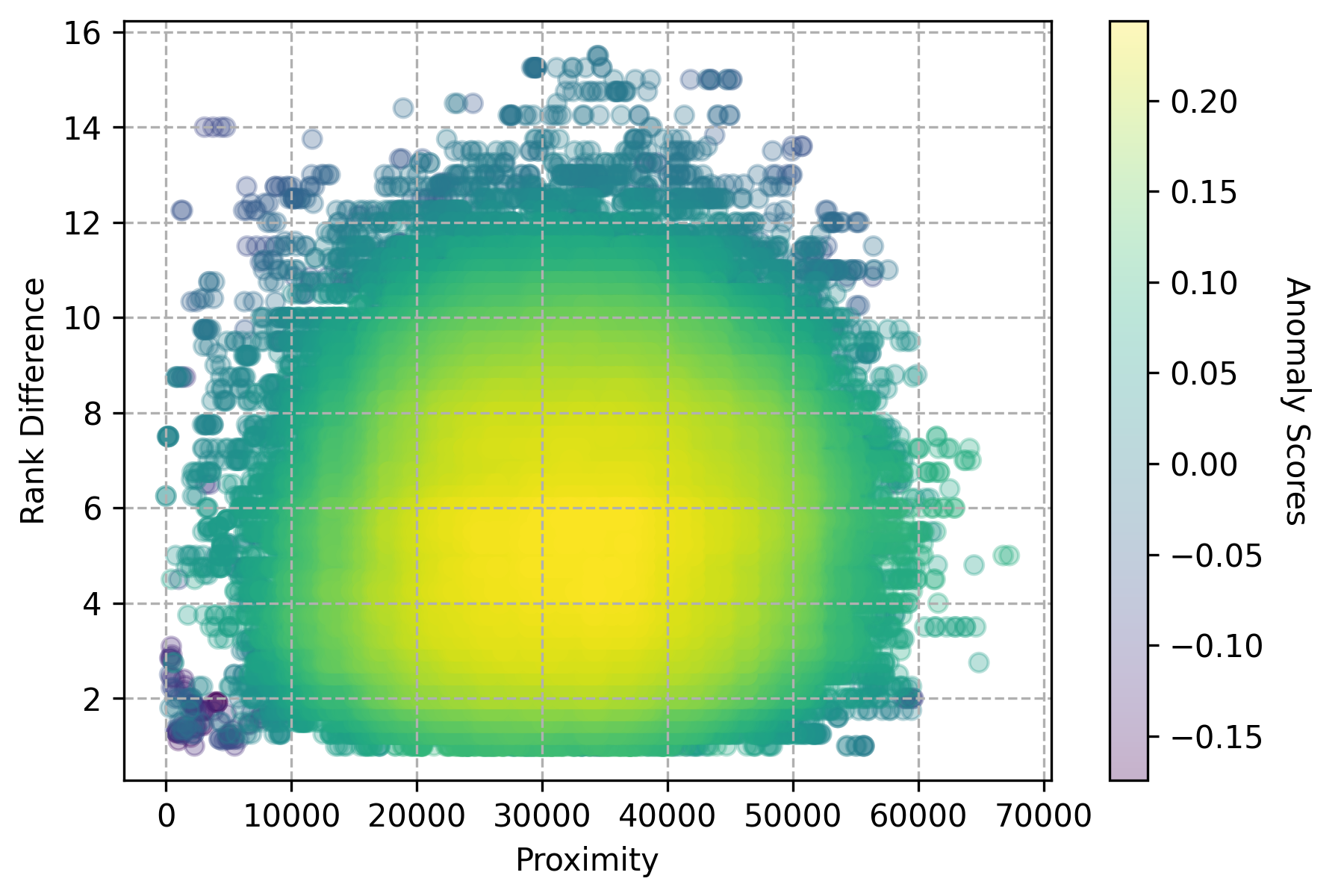}
\caption{Dataset 1}
\label{fig:d1}
\end{subfigure}
\begin{subfigure}[t]{0.49\textwidth}
\includegraphics[width=\textwidth]{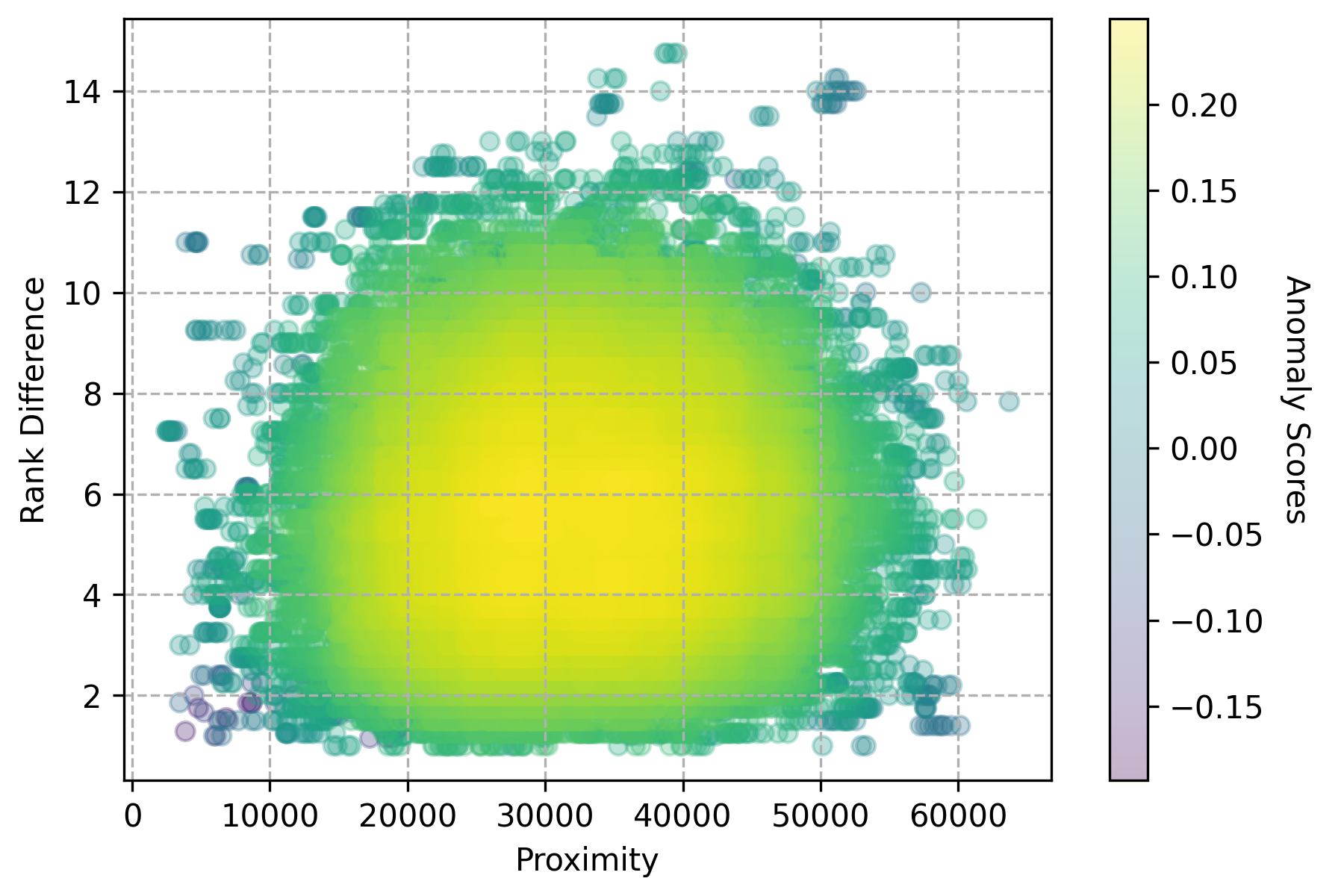}
\caption{Dataset 2}
\label{fig:d2}
\end{subfigure}
\caption{Visual representation of each pair of opponent's rank difference with regards to their average proximity to one another. A darker color represents a more abnormal behavior between the opponents.}
\label{fig:datasets}
\end{figure}

\begin{figure}[h]
\centering
\begin{subfigure}[t]{0.49\textwidth}
\includegraphics[width=\textwidth]{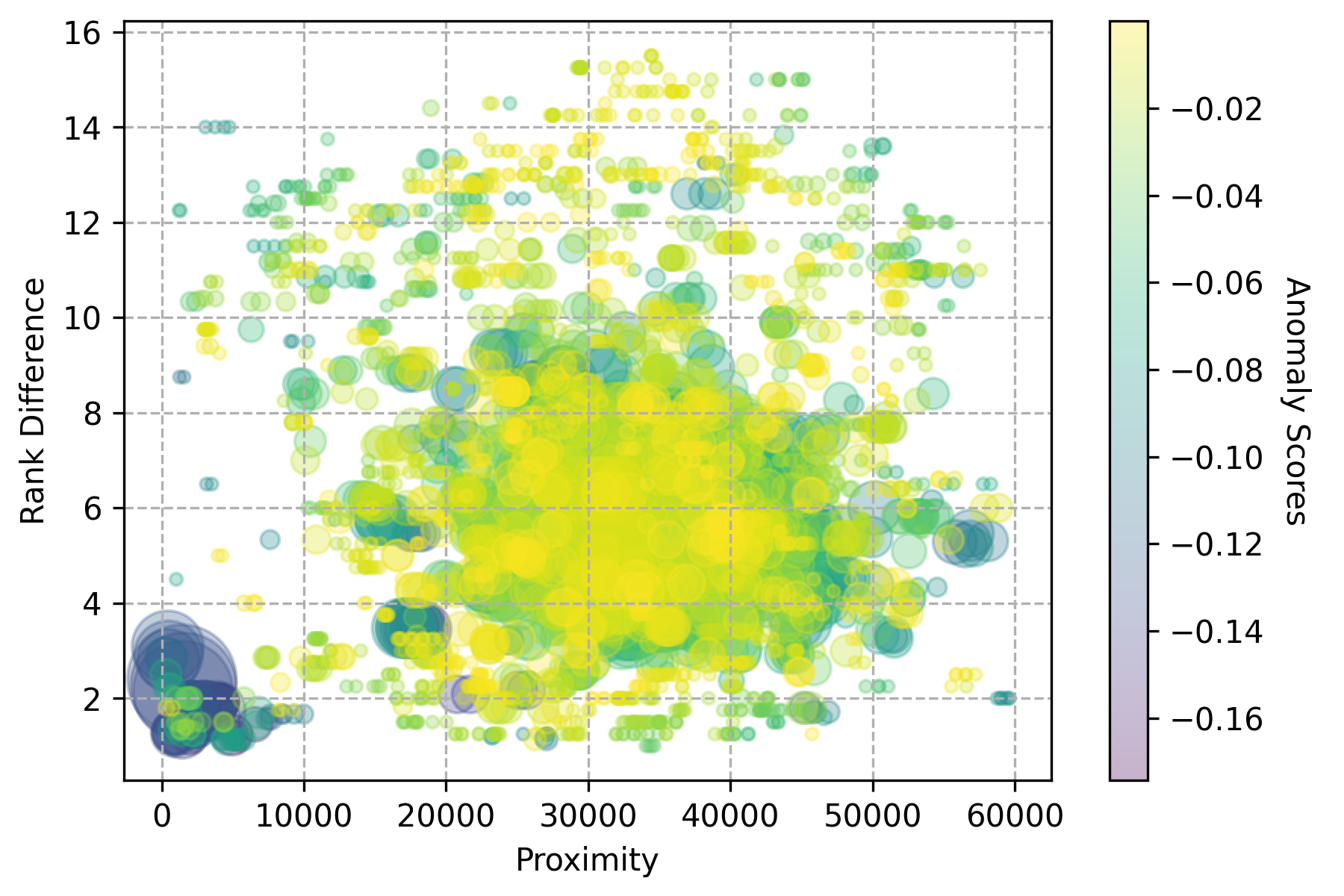}
\includegraphics[width=\textwidth]{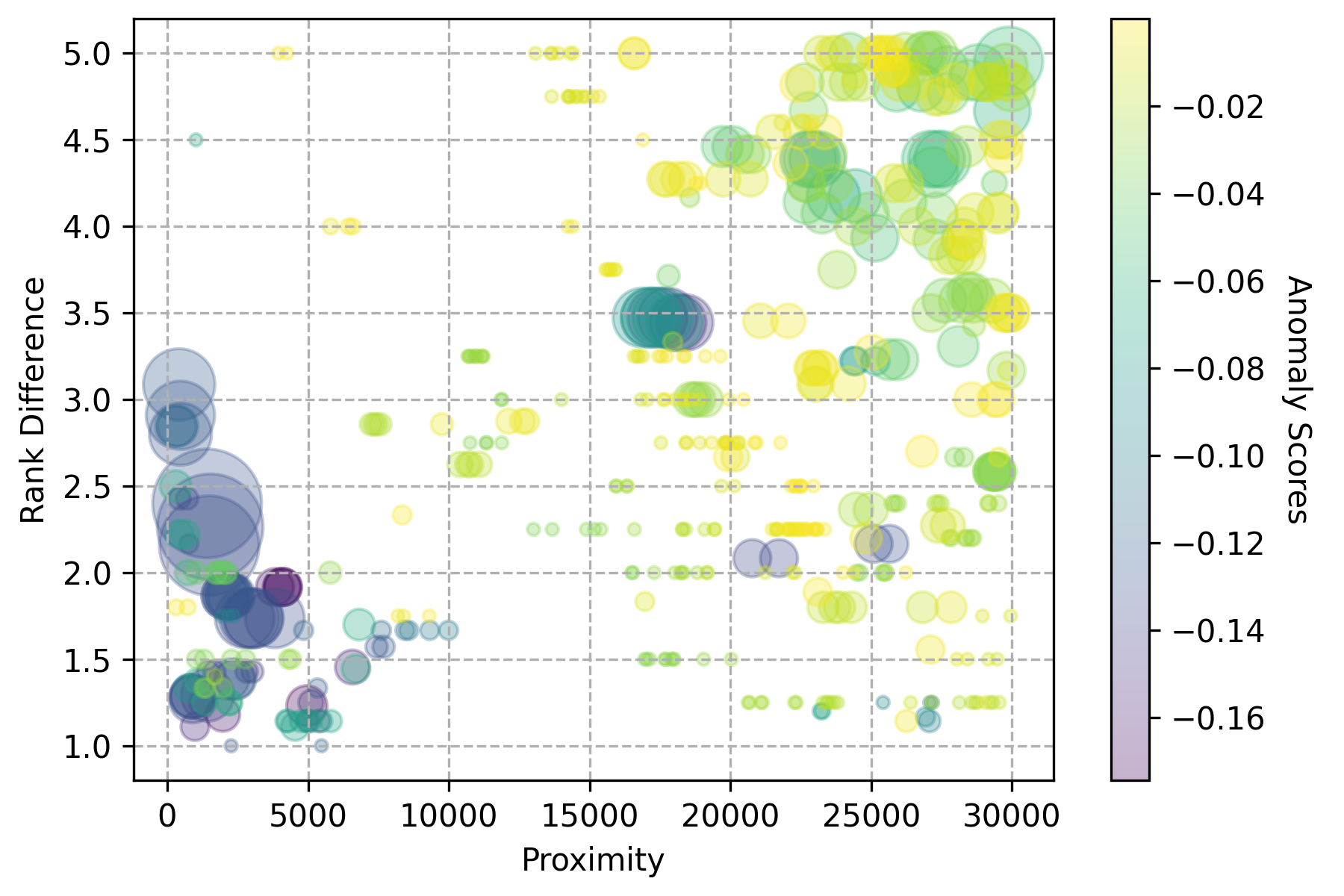}
\caption{Dataset 1}
\label{fig:outliers-d1}
\end{subfigure}
\begin{subfigure}[t]{0.49\textwidth}
\includegraphics[width=\textwidth]{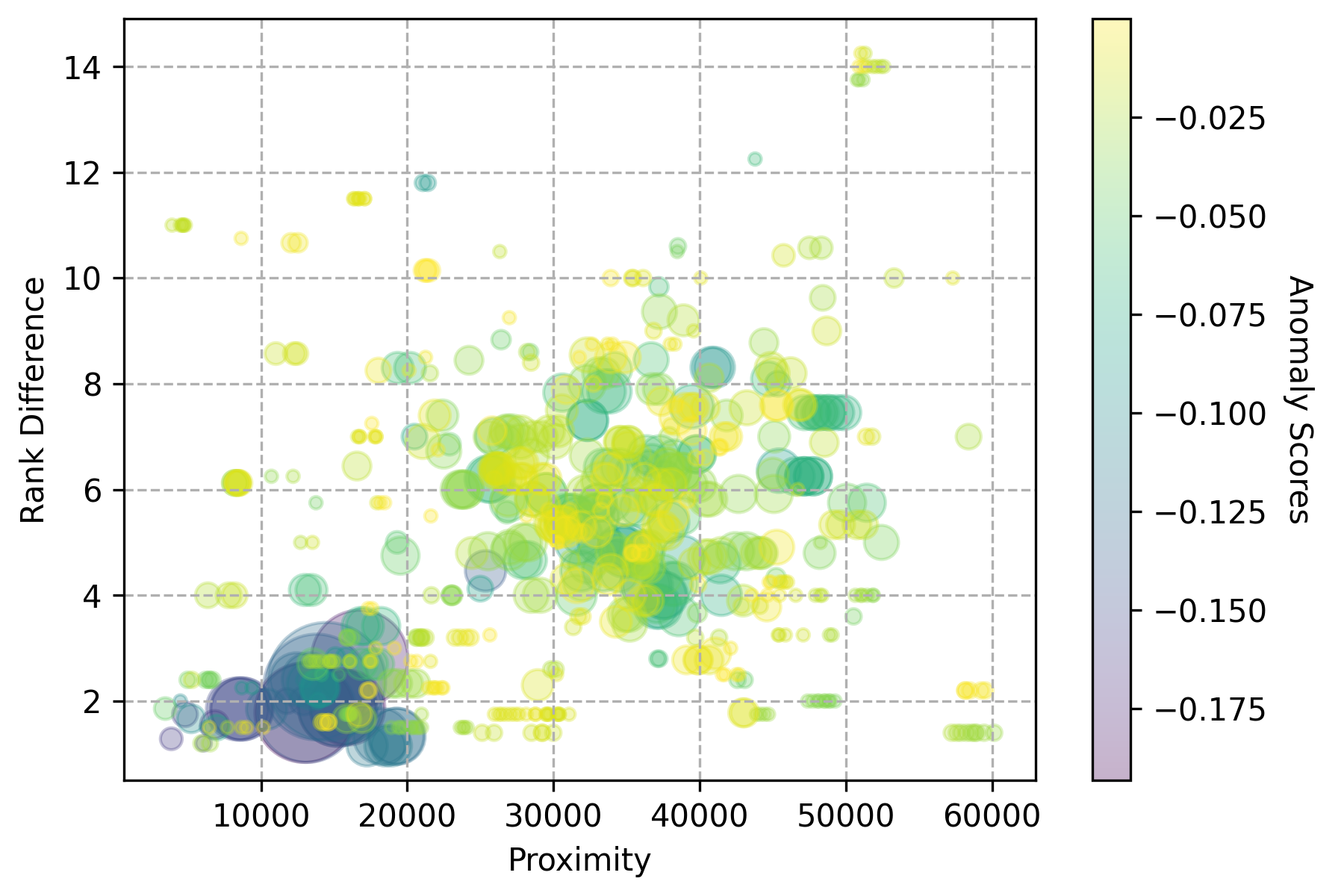}
\includegraphics[width=\textwidth]{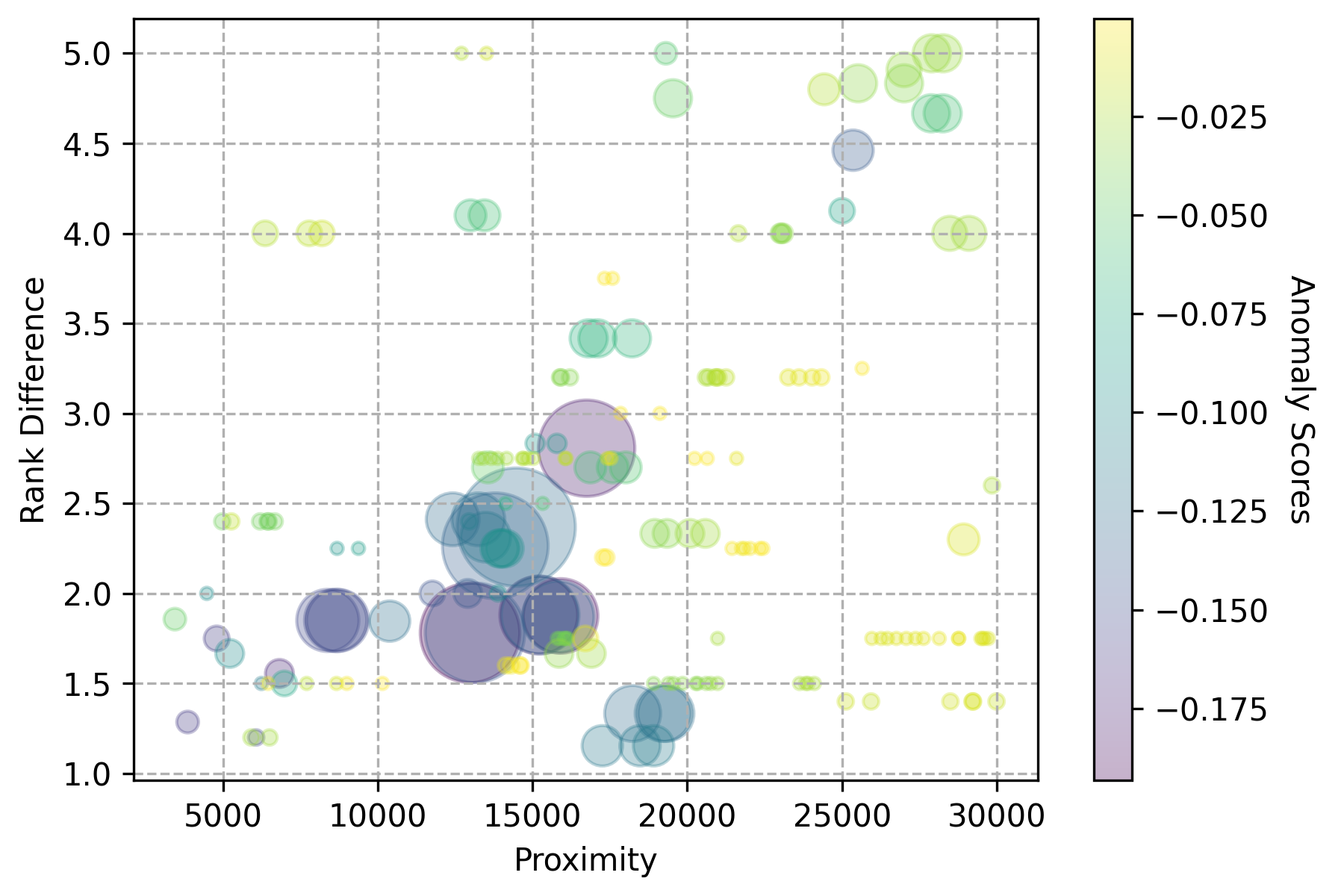}
\caption{Dataset 2}
\label{fig:outliers-d2}
\end{subfigure}
\caption{Visual representation of the flagged outliers by the isolation forest  for both datasets. A darker color represents a more abnormal behavior between the pair of opponents. Sizes represent the number of matches played. On the bottom figures, we show a zoomed in version of the scatter plots for the cluster of outliers in the bottom left corner, i.e. pairs of opponents with a lower average rank placement difference and closer proximity.}
\label{fig:outliers}
\end{figure}

In Figures \ref{fig:d1} and \ref{fig:d2}, we create scatter plots for both datasets such that each dot represents a pair of opponents. Each pair of opponents is given an anomaly score which indicates the degree of collusion exhibited such that negative scores represent outliers and the lower the score, the more abnormal their behavior is. In both datasets, the pairs of opponents that exhibit the highest degree of collusion are clustered in the bottom left corner of the plots as shown in Figures \ref{fig:outliers-d1} and \ref{fig:outliers-d2}, where both the rank difference and proximity are low. We confirm that all suspected colluders mentioned in previous sections as well as the confirmed colluders from Dataset 2 were all identified by the isolation forest and a more detailed look into their behavior is reported in the next subsection. While some pairs of opponents are flagged as outliers, they are not necessarily colluders. In Dataset 2, the isolation forest identifies the pairs with a rank difference higher than 14 as outliers (cf. Figure \ref{fig:outliers-d2}), and while that is true, they do not exhibit any of the features mentioned earlier a part from being somewhat in close proximity. The biggest feature that differentiates these pairs from the rest is the high average rank difference, which is most likely why they have been identified as outliers. As mentioned in Section \ref{sec:method}, it is important to note that the algorithm does not take any action on the detected outliers, as it simply flags potential colluders. Anomaly scores are provided to the corresponding game team and designers to facilitate the detection of collusion and further investigation would be required by human experts before taking any action against these players.

\subsection{Evaluation}
\label{sub:evaluation}

Collusion is still a relatively recent behavior emerging in team-based multiplayer games and with the lack of a training set, we can manually verify each outlier flagged through our existing knowledge on colluding behaviors. Prior knowledge characterizes colluding behaviors similar to ones of teammates, i.e. one that indicates a prior established connection by either having played as teammates before or by appearing in the same set of matches as opponents, close proximity to one another and close in terms of final rank placements. Each pair of opponents is assigned an anomaly score which indicates the degree of collusion exhibited by the pair of players, such that negative scores represent outliers and the lower the score, the more abnormal the behavior.

Out of the 288 outliers detected in Dataset 1, 42 were acquaintances. In particular, out of the 20 pairs of opponents with the lowest anomaly scores, 15 of them had previously played as teammates. Moreover, 78 pairs of opponents flagged as outliers played over 10 matches on opposing teams, with 42.3\% of them averaging a rank difference of less than 3 ranks. Looking specifically at all four features mentioned in this paper, out of the 288 outliers, 72 pairs of players played over 6 games as opponents, had an average proximity of less than 8000 units, and a rank difference of less than 3 over all games played. 65 of those pairs also appeared in the top half of outliers with the lowest anomaly scores. Similarly, out of the 56 outliers detected in Dataset 2, 13 were acquaintances such that 9 of those pairs appear in the top 20 pairs of opponents with the lowest anomaly scores. Although the outliers detected in this dataset did not show as close as an average proximity as the outliers in Dataset 1 (average proximity of all 56 pairs of opponents was 10,000 game units), they did show a stronger indication of prior communication, with 16 pairs of opponents appearing in the same 3 or more consecutive games and 24 of them appearing in over 10 matches on opposing teams. Finally, 42 out of the 56 outliers were less than 3 rank placements apart, averaged over all matches played.

\begin{table}[h]
\def\arraystretch{1.1}
\centering
\resizebox{\columnwidth}{!}{%
\begin{tabular}{|l||c||c|c|c|c|c|c|c|}\hhline{~~-------}
\multicolumn{2}{c|}{} & \textbf{Acquaintance} & \textbf{Rank Difference} & \textbf{Max \# Consec Games} & \textbf{Proximity} & \textbf{\# Matches} & \textbf{Anomaly Score} & \textbf{Colluders} \\\hhline{--=======}
\multirow{5}{*}{\rotatebox{90}{\textbf{Dataset 1}}} & A & TRUE & 1.18 & 2 & 1971.32 & 11 & -0.173 & TRUE\\\hhline{~--------}
    & B & TRUE & 1.11 & 2 & 984.15 & 9 & -0.166 & TRUE\\\hhline{~--------}
    & C & TRUE & 1.45 & 2 & 6573.93 & 11 & -0.162 & TRUE\\\hhline{~--------}
    & D & TRUE & 3.44 & 2 & 18137.61 & 18 & -0.156 & FALSE\\\hhline{~--------}
    & E & TRUE & 3.44 & 2 & 18389.09 & 18 & -0.156 & FALSE\\\hhline{=========}
\multirow{5}{*}{\rotatebox{90}{\textbf{Dataset 2}}} & A & TRUE & 1.78 & 3 & 12982.80 & 32 & -0.184& TRUE\\\hhline{~--------}
    & B & TRUE & 1.88 & 3 & 15201.66 & 25 & -0.157 & TRUE\\\hhline{~--------}
    & C & TRUE & 1.875 & 3 & 15910.2 & 24 & -0.154& TRUE\\\hhline{~--------}
    & D & TRUE & 1.56 & 2 & 6825.69 & 9 & -0.149 & TRUE\\\hhline{~--------}
    & E & TRUE & 1.2 & 2 & 6072.28 & 5 & -0.13 & TRUE\\\hhline{---------}
\end{tabular}}
\caption{Top 5 anomalies with the lowest anomaly scores detected in Dataset 1 and Dataset 2. \textbf{Colluders} column is set to TRUE if the pair of opponents are confirmed to be in fact colluders by the game designers, FALSE otherwise. This is done by manually checking players' gameplay data for each game play during the corresponding date range.}
\label{tab:outlier-stats}
\end{table}

We report the gameplay data for the top 5 pairs of opponents exhibiting the highest degree of collusion in both datasets in Table \ref{tab:outlier-stats} (i.e. pairs with the lowest anomaly scores). We confirm with the game designers that 14 out of the 20 most abnormal pairs from Dataset 1 were most likely to be colluding, and 16 out of the first 20 most abnormal pairs from Dataset 2 were all confirmed colluders. According to the game designers, the biggest factors in deciding whether two teams are in fact colluding are the number of consecutive matches and repeated close final rank placements. The probability for this to occur is very low to be a coincidence over a number of consecutive matches or a high number of matches in such a given short time frame. That said, pairs D and E from Dataset 1, although exhibiting a suspicious behavior by appearing in multiple consecutive matches on a more extended date range, could not be confirmed to be colluders due to their disparate placings throughout the games.

When collusion is clearly occurring, game designers spend approximately 3 minutes manually checking opposing team's gameplay data. In less obvious cases, game designers need to search through more matches (as was the case with pairs D and E from the previous example), which can significantly increase the time required to detect colluding teams. By isolating the outliers in our dataset, we help game designers reduce the pool of players needed to be manually investigated, focalizing only on the players exhibiting the highest degree of collusion in our dataset.

\section{Concluding Remarks}
\label{sec:conclusion}

While it may not be feasible to entirely eliminate cheating and collusion from games, they can be detected and the players involved can be punished accordingly. In this paper, we focused on team-based multiplayer games where collusion happens between teams of players. Using a player’s external and in-game social relationships paired with its in-game behavioral patterns, we are able to infer cross-team collusion in games. We use isolation forests to automate the detection and assign each pair of opponents a collusion score. We note that this approach does not determine whether two teams are in fact colluding, but rather indicates the degree of collusion exhibited by the pair's respective teams. Hence, human intervention is still essential and further investigation by human experts is required before taking any enforcement action against the potential colluders flagged by our model. We intend on pursuing this work to better automate the detection of collusion in team-based multiplayer games by analyzing the evolution of a players' behavior over larger datasets by relaxing some constraints such as the number of matches played, and on longer time periods. Furthermore, in our analysis, we concentrate on team proximity at the start of a match. We would like to expand this to consider in-game team positions at different time steps during a match as this will provide more information regarding team proximity and their relations to opposing teams.

\small
\bibliography{refs}

\end{document}